\documentclass[letterpaper, 10 pt, conference]{ieeeconf}  % Comment this line out if you need a4paper
\IEEEoverridecommandlockouts  
\usepackage{multicol}
\usepackage{xcolor}
\usepackage{mathtools}

\usepackage{hyperref}
\usepackage{graphics,graphicx,caption,float,subcaption,booktabs,xcolor,multirow,array,color,ifthen,tabu,colortbl,dblfloatfix,url,xparse,mathtools,patchcmd,algorithmic,amssymb,xspace,nicefrac,microtype,amsmath,amsfonts,bm,ragged2e,tikz,stackengine,etoolbox,xpatch,enumerate,xstring,setspace,tabularx,makecell,changepage,cuted,titlesec,wrapfig,tcolorbox,footmisc, xspace}

\newcommand\raven[1]{\textcolor{orange}{}}

\title{\LARGE \bf
Manipulator as a Tail:\\ Promoting Dynamic Stability for Legged Locomotion}

\author{Huang Huang$^{1}$, Antonio Loquercio$^{1}$, Ashish Kumar, Neerja Thakkar$^{1}$, Ken Goldberg$^{1}$, Jitendra Malik$^{1}$ \thanks{$^{1}$ University of California, Berkeley}%
}

\begin{document}

\maketitle
\thispagestyle{empty}
\pagestyle{empty}

\begin{abstract}
%Deployment of a legged system with a mounted arm requires controllers that can jointly control the legs and arm to use the available limbs optimally for mobile manipulation tasks.
For locomotion, is an arm on a legged robot a liability or an asset for locomotion? 
%Prior works mainly designed specific controllers to account for the added payload and inertia from a fixed manipulator. 
Biological systems evolved additional limbs beyond legs that facilitates postural control.
% typically benefit from additional limbs, which can simplify postural control. 
% For instance, 
% geckos use their tails to enhance the stability of their bodies and prevent falls under disturbances and cheetahs use their tails to turn at high speeds when running.
This work shows how a manipulator can be an asset for legged locomotion at high speeds or under external perturbations, where the arm serves beyond manipulation. Since the system has 15 degrees of freedom (twelve for the legged robot and three for the arm), off-the-shelf reinforcement learning (RL) algorithms struggle to learn effective locomotion policies.
Inspired by Bernstein's neurophysiological theory of animal motor learning, we develop an incremental training procedure that initially freezes some degrees of freedom and gradually releases them, using behaviour cloning (BC) from an early learning procedure to guide optimization in later learning. 
% We name our algorithm RBIL (\textbf{R}L and \textbf{B}C with \textbf{I}ncremental \textbf{L}earning).
% This philosophy of stage-wise training is reminiscent of Nikolai Bernstein’s ~\cite{bernstein1966co,bernstein2014dexterity} proposal for biological motor learning by first freezing some degrees of freedom to make the task easier, and after it has been mastered in that framework, by unfreezing them to improve performance further.
%
Simulation experiments show that our policy increases the success rate by up to 61 percentage points over the baselines.
Simulation and real robot experiments suggest that our policy learns to use the arm as a ``tail" to initiate robot turning at high speeds and to stabilize the quadruped under external perturbations. Quantitatively, in simulation experiments, we cut the failure rate up to 43.6\% during high-speed turning and up to 31.8\% for quadruped under external forces compared to using a locked arm.

% with up to 13\% and 32\% success rate percentage improvement compared to using a locked arm in simulation turning and perturbation experiments respectively. 
%Our findings show the role of ``tail'' movements in legged robots, resembling the role it plays for other terrestrial vertebrates.

% To do so, we train a sensorimotor policy using deep reinforcement learning to create a synergy between the robot's limbs. 
% This policy enables the robot to maintain stability despite large disturbances.
% %
% However, learning such a controller can be quite challenging.
% %
% To account for these challenges, we propose a stage-wise training procedure to learn complex behaviors. 
%
%Chimera achieves a 15\% higher success rate compared to baseline methods. We deploy our policy in the real world and show qualitative examples of our reactive controller that controls the arm and the legs to achieve stability under strong disturbances.         
%we propose a method, Chimera, to learn controllers capable of carrying out locomotion against external disturbances with a complex system such as an A1 quadruped robot with a mounted WidowX 200 arm. 
% Our proposed method decomposes this complex task into three stages and then incrementally learns these tasks to arrive at a single policy capable of solving the final control task, achieving a success rate up to 2.35 times higher than baselines in simulation. We deploy our learned policy in the real world and show stability during locomotion under strong disturbances.            
\end{abstract}
% \keywords{Quadruped Locomotion; Arm on a Quadruped; Curriculum learning;} 

% \setcounter{figure}{1}  
% \IEEEpeerreviewmaketitle

\section{Introduction} 
\label{sec:introduction}

% \begin{itemize}
%     \item Biologically, there are many examples of arms/tails being helpful for humans and animals to balance when met with external forces or navigating on difficult terrain
%     \item Most prior work on mobile manipulation with legged robots has focused on combining manipulation and the locomotion
%     \item Here, we use a manipulator to improve the ability of our robot dog on hard terrain, under external forces
% \end{itemize}
Human arms are vital for manipulation, but also improve stability. For instance, humans swing their arms to balance when falling or when being pushed~\cite{ting2007neuromechanics}. Similarly, many animals use their tails to improve their stability and agility. A beaver's wide, flat tail helps it balance its body weight and skillfully maneuver through the water. 
A cheetah's tail helps it make sharp turns at astonishing speeds \cite{patel2013rapid}. 
% Human arms and hands are vital for our ability to skillfully manipulate objects. However, the utility of our limbs goes far beyond our manipulation abilities. 
% % When children learn how to climb up and down stairs, they cling to the railing or use their arms for balance. 
% When we suddenly fall, we instinctively throw out an arm to catch ourselves. If someone pushes us, we use our arms to counter the force~\cite{ting2007neuromechanics}. Similarly, many animals use their tails to improve stability and agility when navigating difficult terrain. A beaver's wide, flat tail helps it balance its bodyweight and skillfully maneuver through the water. 
% % A squirrel's success at staying balanced on even the most minute tree branches hinges on the counterbalance provided by its large tail. 
% A cheetah's tail helps it make sharp turns at an astonishing speed. Overall, animal tails have a wide variety of uses -- they can assist with communication, escaping predators, and prehensile tails can additionally assist with grasping and manipulation of objects~\cite{dickinson2000animals,holmes2006dynamics}. 

% Recent works have studied adding an arm to legged robots for mobile manipulation~\cite{fu2022deep,sleiman2021unified,bellicoso2019alma,zimmermann2021go,chiu2022collision}. The attached manipulator enlarges the robot capability, it brings challenges to the robot stability. Prior works focus on maintaining robustness while the arm performs manipulation tasks~\cite{sleiman2021unified,bellicoso2019alma,zimmermann2021go,chiu2022collision}. We 

Prior works~\cite{fu2022deep,sleiman2021unified,bellicoso2019alma,zimmermann2021go,chiu2022collision,kumar2024practice} have added an arm to legged robots for mobile manipulation and focus on maintaining robustness while the arm performs manipulation tasks.
Recent model-based approaches propose whole-body planners that use robotic arms beyond manipulation.
By formulating a multi-contact optimal control problem, they show that an arm can help the balance of the robot~\cite{sleiman2021unified,chiu2022collision, sleiman2023versatile}.
%
%This is achieved by formulating a multi-contact optimal control problem~\cite{chiu2022collision,sleiman2021unified}.
However, these approaches are sensitive to the fidelity of the model used for planning~\cite{sleiman2021unified}, require specific self-collision routines~\cite{chiu2022collision}, and need accurate state estimators (e.g., linear velocity) for trajectory tracking~\cite{sleiman2023versatile}.
These requirements constrain these approaches to relatively large robots carrying enough sensors and computing for the required computation.

Inspired by biology, we explore how a low-cost and small robot-manipulator setup can learn to actively use the arm to benefit the stability and agility of locomotion under external perturbations and at high speeds.
%
% controlled arm can benefit the stability and agility of locomotion under external perturbations and at high speeds.
Using an arm to balance or turn is a complex whole-body control problem requiring synergy between all the robot limbs.
It is a highly dynamic task: the robot must move its limbs quickly to generate sufficient torques to balance against external perturbations or enough momentum to turn fast. 

% One traditional approach to this problem consists of whole-body planning by formulating a multi-contact optimal control problem~\cite{chiu2022collision,sleiman2021unified}, which are sensitive to the fidelity of the model used for planning and require accurate state estimators for control.
% %
% Online optimization of a whole-body control problem can also be computationally very expensive.
% %
% Therefore, these methods have been constrained to relatively large robots which can carry enough sensors and compute.
\begin{figure}
    \centering
    \includegraphics[width=\linewidth]{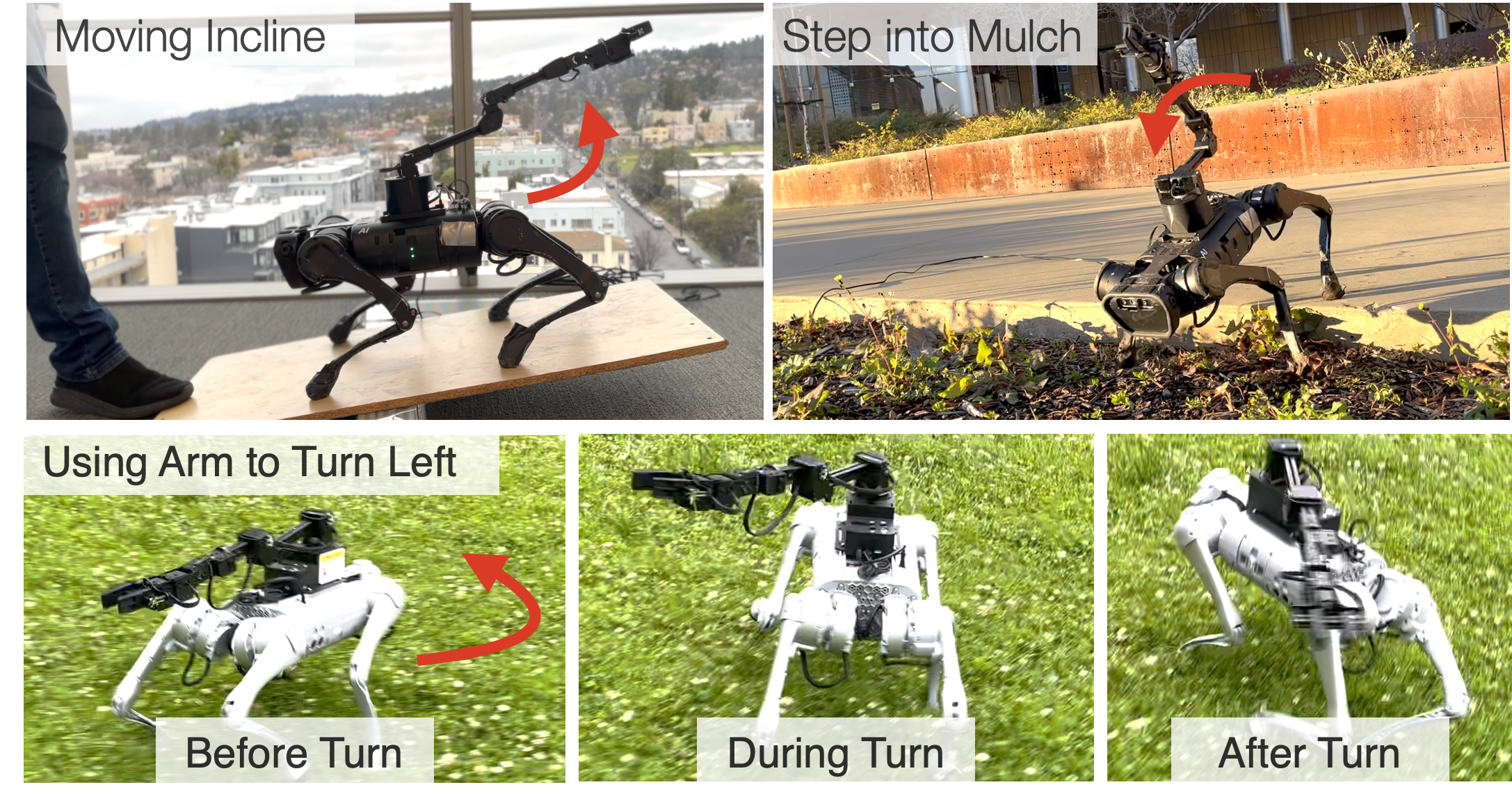}
    \captionof{figure}{A learned whole-body control policy increases a legged robot's stability and agility. We evaluate the dynamic benefits of the arm in two scenarios. \textbf{Top}: stabilization under dynamic external forces. The actuated arm responds to an impulse force by swinging quickly away from the impulse direction to generate a balancing impulse force. As shown in section~\ref{sec:math}, the arm relies on such dynamic forces instead of the static CoM changes to stablize.
 % \textbf{Middle:} stabilization on a moving incline. The robot arm moves upwards to mitigate the instant torque and settles to the leveled position when the incline is leveled as in the last frame. 
 \textbf{Bottom:} dynamic agile locomotion, where the arm helps the quadruped turn left at high speeds. Before and during the turn, the arm moves toward the turning direction to increase the centripetal acceleration (toward the turning direction), similar to behaviors observed in animals. After the turn, the arm moves back to the nominal position.
 %The first row shows the robot arm moving towards the pushing direction to generate the opposite torque on the robot base to balance again a human kick. From left to right, the frame shows the arm position before kick, arm moving during kick and arm moving back after the kick. The second row shows the robot arm moving up and down to balance on the moving incline. When the incline moves as in the second frame, the robot arm moves upwards to mitigate the instant torque and settles to the leveled position when the incline is leveled as in the last frame. 
 Videos and supplementary materials are here: \url{https://tinyurl.com/2p8edezu}
 % \href{https://tinyurl.com/2p8edezu}{here}.
 }
 \vspace{-20pt}
    \label{fig:fig1}
\end{figure}

Learning-based methods have shown success for legged locomotion~\cite{kumar2021rma,loquercio2022learning,agarwal2022ego,lee2020learning,miki2022learning}. We learn a unified sensorimotor policy for both the arm and the quadruped.
% \textit{on the task of walking}.
%
%In contrast to previous work, we do not train the manipulator for a specific reaching task~\cite{fu2022deep,zimmermann2021go}, with the legged robot compensating for changes 
% However, orchestrating the degrees of freedom of the manipulator and the legged robot is a complex optimization problem.
To simplify the complex optimization problem for a  system with many degrees of freedom, we use an incremental stage-wise training procedure that gradually increases the number of controllable degrees of freedom.
% inspired by Nikolai Bernstein’s ~\cite{bernstein1966co,bernstein2014dexterity} proposal for biological motor learning.
%
Specifically, we learn incrementally more complex behavior with RL,
using BC from previously learned behaviors as a proxy loss to guide the optimization.
%
% This philosophy of stage-wise training is reminiscent of Nikolai Bernstein’s ~\cite{bernstein1966co,bernstein2014dexterity} proposal for biological motor learning by first freezing some degrees of freedom to make the task easier, and after it has been mastered in that framework, by unfreezing them to improve performance further.
Using demonstrations and incremental learning to improve RL-based controllers has been widely explored for locomotion~\cite{hester2018deep,goecks2019integrating,peng2018deepmimic,peng2020learning,peng1994incremental}.
In this paper, we show the unification of these two concepts is particularly suitable for our problem setting.
Empirically, we show up to 30 percentage points higher success rate in walking compared to using demonstrations without incremental learning and up to a 50 percentage points higher success rate than incremental learning without demonstrations (i.e., curriculum learning), as shown in Table~\ref{tab:results_set2}.

We deploy the learned policy on a physical robot in the real world and evaluate the emerging locomotion strategy in both simulation and real-world settings for two locomotion tasks: stable locomotion to balance against external perturbations and agile locomotion to turn at high speed.
We observe that the actuated arm compensates against pushes and stabilizes under external forces and on a moving incline (Fig.~\ref{fig:fig1}).
We also find that the learned policy exhibits behaviors that are reminiscent of the motion of geckos~\cite{Siddall2021mechanisms}:
when turning at high speeds ($\approx 1.5$ m/s), the tail precedes the motion of the body to increase centripetal forces (Fig.~\ref{fig:fig1}).
In addition, we interpret our results in the light of a first-principle analysis. The analysis shows that while the arm is not useful statically due to its small mass, it clearly benefits the dynamic stability.

\section{Related Work}

% \subsection{Animals and Robots with Tails}
Many biologists and roboticists have studied the utility of animal tails in maneuverability and stability~\cite{Shield2021tails}. For instance, cats and  squirrels use their tails to facilitate balancing~\cite{walker1998balance, fukushima2021inertial}. Lizards and cheetahs use their tails to effectively reorient their bodies at high speeds~\cite{Siddall2021mechanisms,patel2013rapid}. 
These utilities of tails helped the first terrestrial organisms adapt from aquatic environments to various terrestrial ones, by improving their stability and agility on difficult terrains~\cite{mcinroe2016tail}.
% Many biologists and roboticists have studied the utility of animal tails in maneuverability and stability~\cite{Shield2021tails}. For instance, a study demonstrated that when cats traversed narrow, moving beams, their tails facilitated balance by helping keep their hips aligned over the beam~\cite{walker1998balance}. Preventing cat tail usage resulted in a significantly higher incidence of falls. When squirrels fall, their tail is critical in enabling them to land gracefully upright and avoid injuring themselves~\cite{fukushima2021inertial}, and lizards also exhibit this phenomena, with intertial forces from their tail being more effective at reorienting them than aerodynamic forces~\cite{Siddall2021mechanisms}. Cheetahs use their tails to provide a torque that counteracts the centrifugal force of fast turns, enabling them to stay balanced through rapid, high-speed turns~\cite{patel2013rapid}. These utilities of tails have likely existed since very early times, significantly helping the first terrestrial organisms adapt from aquatic environments to various terrestrial ones, by improving their stability in moving across difficult terrains \cite{mcinroe2016tail}. 

Tails have also proven beneficial in robotic design. An attached tail can assist a robot in flipping upright and achieving a balanced landing while in free fall~\cite{johnson2012tail,chang2013nonlinear,fukushima2021inertial} or when jumping~\cite{zhao2013controlling,an2020development}. 
% As in animals, tail and head movements impact the locomotion of a robot~\cite{zhang2016effects}, and t
Tails can help robots maneuver through turns at higher speeds~\cite{patel2013rapid} and navigate difficult terrain that would otherwise trap the robot~\cite{soto2022simplifying}.
The Arque tail helps humans balance by counteracting changes in the wearer's center of mass and momentum~\cite{Nabeshima2019}.
% The Arque tail is a wearable robotic tail for humans that assists with balance by counteracting changes in the wearer's center of mass and momentum~\cite{Nabeshima2019}.
Inspired by the utility of tails in biology and robotics, we use a manipulator as a tail to enhance locomotion stability and agility. A few other works have also used limbs to counteract disturbances, in similar function to a tail. For instance, \cite{khazoom2022humanoid} proposes a model that plans arm motions in reaction to forces on a humanoid robot, allowing its center of mass to stay within the support polygon. \cite{ferrolho2022RoLoMa} focus on robust locomotion in a quadruped robot with an arm, keeping the base stable as the arm carries out a manipulation task. In~\cite{ferrolho2020optimizing}, a standing quadruped robot with an arm attached is disturbed, and an arm trajectory helps counteract forces.~\cite{ma2023learning} learns a policy to use the arm to prevent the quadruped from falling or help the quadruped to recover from falling. In contrast, our method counteracts forces with dynamic arm movements while the quadruped is moving and uses arm to enhance quadruped agility.

 Some prior works focus on mobile manipulation and train the arm separately from the robot base, and combine the two~\cite{Zimmermann2021GoF,ma2022combining,quesada2022holo}. Other work involves jointly learning locomotion and manipulation on the whole body, reducing engineering and tuning needed to combine the two components and allowing for more complex behaviour to emerge~\cite{fu2022deep,sleiman2021unified}. Our approach similarly jointly learns the behaviour of the robot body and the arm, but carries out this learning in multiple stages to simplify the optimization problem. We show another use of the manipulator beyond manipulation to benefit locomotion, in contrast to prior works~\cite{ma2022combining,ferrolho2020optimizing} that consider the arm as a liability for locomotion.

%\subsection{Combining Behavior Cloning and Reinforcement Learning}

% \paragraph{Combining Behavior Cloning and Reinforcement Learning} 
Using demonstrations to improve the training process of RL algorithms has a long history~\cite{schaal1996learning,atkeson1997robot,hester2018deep,goecks2019integrating,peng2018deepmimic,peng2020learning,xie20learning}.
%
% More recently, the idea has been applied in the context of deep RL~\cite{hester2018deep,goecks2019integrating,peng2018deepmimic,peng2020learning}. 
These works find that when expert demonstrations are available, training policies with RL is faster and more effective. Additionally, incremental RL learning has proven to be successful~\cite{wang2019incremental, wang2021lifelong}.
%
% We combine BC from demonstrations with RL using an annealing weight to achieve multi-stage incremental learning of whole-body control.
We made the following key changes to achieve the unification of learning from demonstrations, RL and incremental learning  to achieve multi-stage incremental learning of whole-body control. We incrementally learn policies trained on simpler skills to provide demonstrations for policies trained on more complex ones. Policies from early stages provide on-policy demonstrations using an annealing weight instead of static external demonstrations for later stage, which is shown to favor training~\cite{ross2011reduction, loquercio2021learning, loquercio2019deep}. 

\section{Method} 
\label{sec:method}

\begin{figure}[t]
    \centering
    \vspace{3pt}
    \includegraphics[width=0.48\textwidth]{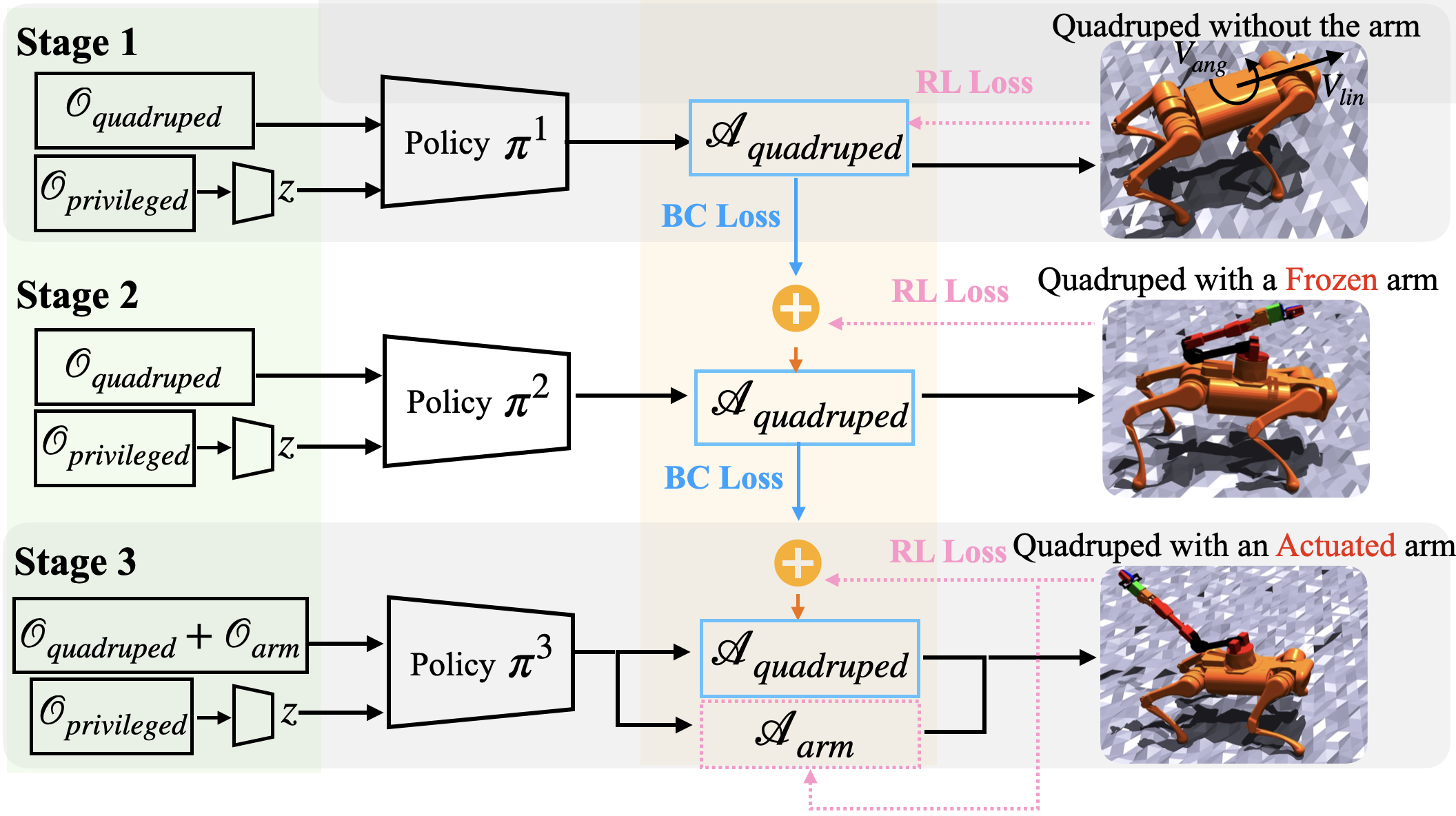}
    \caption{
    We learn whole-body control policies for quadruped locomotion with an arm in three stages, each learning a different skill of increasing complexity. Policies at each stage observe $\mathcal{O}_{quadruped}$ or $\mathcal{O}_{arm}$ and an extrinsic vector $z$ with compressed information of privileged observations $\mathcal{O}_{privileged}$, which consists of mass, friction, quadruped base velocity and external perturbations.
    }
    %$\pi^{2}$ is used for calculating the BC losses on the quadruped. The RL loss is applied for both the quadruped and the arm. In addition to $\mathcal{O}_{quadruped}$ and $\mathcal{O}_{privilege}$, $\pi^3$ also observes $\mathcal{O}_{arm}$, which includes the arm joint positions, velocities and previous actions. $\pi^3$ outputs both $\mathcal{A}_{quadruped}$ and $\mathcal{A}_{arm}$, which are the target positions of the arm. }
    \label{fig:3-stage} 
    \vspace{-20pt}
    \centering

\end{figure}

We use an iterative optimization algorithm for synthesizing behaviors for a task $\mathcal{K}$. We assume that the task $\mathcal{K}$ can be decomposed into multiple stages: $\mathcal{K}^1$...$\mathcal{K}^n$~\cite{singh1992transfer,dean1995decomposition,dietterich2000hierarchical}.
The stage $\mathcal{K}^i = (\mathcal{O}^i, \mathcal{R}^i, \mathcal{A}^i)$ has $\mathcal{O}^i$ as the observation space, $\mathcal{R}^i$ as the reward function, and $\mathcal{A}^i$ as the action space. Each stage $\mathcal{K}^i$ includes all the previous stages $\mathcal{K}^{1:i-1}$, along with additional objectives at stage $i$. These objectives could potentially add a term to reward function for obtaining $\mathcal{R}^i$, or involve additional sensors or actuators to get $\mathcal{O}^i$ and $\mathcal{A}^i$ respectively. The training scheme works as follows: \\
\textbf{Base step} ($\mathcal{K}^1$): We train the policy $\pi^1(o^1)$ to predict $a^1$ from scratch using RL to optimize the returns defined by the reward function $\mathcal{R}^1$. \\
\textbf{Inductive step} ($\mathcal{K}^i$): Once we have policy $\pi^{i-1}$, we train policy $\pi^{i}$ by optimizing:
    \begin{equation}
    \begin{aligned}
     \max_{\pi^i} \quad & (J(\pi^i) =\mathbb{E}_{\tau \sim p(\tau|\pi^i)}\Bigg[\sum_{t=0}^{T-1}\gamma^t r^i_t\Bigg])\quad \\
    & \textrm{s.t.} \quad  \mathcal{DW}(\pi^i, \pi^{i-1}) \leq \delta \\
    \end{aligned}
    \end{equation}    
    where $\mathcal{DW}$ is a Wasserstein distance function, $\tau$ is the trajectory of the agent when executing policy $\pi^i$, and $p(\tau|\pi^i)$ represents the likelihood of the trajectory under $\pi^i$.This optimization is similar to the one proposed in~\cite{xie20learning}. In constrast to \cite{xie20learning}, we computes the expectation over the visited states $\tau$ instead of a fixed dataset $\mathcal{D}$ and do not require any initial reference motions. Using the Kantorovich-Rubinstein theorem~\cite{gibbs2002choosing}, we simplify it to a relaxed unconstrained optimization problem  (details on website): 
    \begin{equation}
       \mathcal{L}(\pi^i) = J(\pi^i) + \lambda*\mathbb{E}_{\tau} [ \|a^i_t - a^{i-1}_t\| ],
       \label{eq:final}
    \end{equation}
    where $a^i_t = \pi^i(o^i_t)$, $a^{i-1}_t = \pi^{i-1}(o^i_t)$ (computed over the relevant part of $o^i_t$), and $\|a^i_t - a^{i-1}_t\|$ is a regression loss (computed over compatible part of actions) that encourages $\pi^i$ to stay close to $\pi^{i-1}$.  Note that Eq.~\eqref{eq:final} can be algorithmically optimized by simultaneously minimizing $J(\pi^i)$ using RL and $\mathbb{E}_{\tau} \|a^i_t - a^{i-1}_t\|$ with BC between $\pi^i$ and $\pi^{i-1}$. We anneal $\lambda$ to decrease the imitation constraint in the later stages of training and find an overall better solution. Figure~\ref{fig:3-stage} shows an overview of our approach.
    % \[
    %   \mathcal{L} =  (1-\alpha) * RL(\pi^i) + \alpha * BC(a^i_t, a^{i-1}_t)
    % \]
    % where $a^i_t = \pi^i(o^i_t)$, $a^{i-1}_t = \pi^{i-1}(o^i_t)$ (computed over the relevant part of $o^i_t$), $BC(a^i_t, a^{i-1}_t)$ is a regression loss computed over the compatible part of the actions and RL maximizes the expected return of the policy $\pi^i$: $J(\pi^i)$.
% \[
%     J(\pi^i) = \mathbb{E}_{\tau \sim p(\tau|\pi^i)}\Bigg[\sum_{t=0}^{T-1}\gamma^t r^i_t\Bigg],
% \]
% where $\tau = \{(o^i_0, a^i_0, r^i_0), (x^i_1, a^i_1, r^i_1) . . .\}$ is the trajectory of the agent when executing policy $\pi^i$, and $p(\tau|\pi^i)$ represents the likelihood of the trajectory under $\pi^i$. 
%
% Note that the $RL$ loss can potentially modify behaviors from the previous stages to achieve optimal performance under $\mathcal{R}^i$.

% \subsection{Incremental Stage-wise Learning for Whole-Body Control}

We learn locomotion policies by dividing the task into a three-stage process (Fig.~\ref{fig:3-stage}).
We first train a walking policy $\pi^1$ for locomotion on flat terrain without an arm attached.
Then, we train a policy $\pi^2$ for locomotion in the presence of task specific conditions, where the arm is attached but locked.
Finally, we train a policy $\pi^3$ that controls both the legged robot and the arm under task specific conditions.
The final policy can utilize the arm to balance under strong external pushes or rough terrain and help the quadruped turn at high speed.
For both $\pi^2$ and $\pi^3$, we use BC on the policy from the previous iteration as a proxy loss during optimization. The task-specific conditions in stages 2 and 3 are external perturbations for stable locomotion and turning at high speed for agile locomotion.
%
% At each stage, our approach generates a policy with a higher success rate than its predecessor.
%
% The final policy $\pi^3$ achieves a success rate twice as high as a policy that controls all degrees of freedom from the beginning of the training.

%
% In addition, our approach outperforms a baseline that controls the robot's legs and arms in a decoupled fashion.
% %
% We deploy $\pi^3$ on a physical robot (Fig.~\ref{fig:robot}) in the real world and test the stability of the learned policy in a large set of uneven terrains.
% %
% We find that our approach elicits a synergy between the legs and the arm of the robot, enabling it to compensate for large pushes and stabilize under moving inclines (Fig.~\ref{fig:fig1}).

% \section{Stage-wise L}
% \raven{raisim policy is not trained with external pushes right?}

% We apply our optimization approach to the problem of whole-body control of a quadruped dog with a mounted arm to balance against external perturbations. We divide this task into the following three stages: (a) walking without a mounted arm, (b) walking with a mounted but frozen arm under external perturbations, (c) walking with a controllable arm under external perturbations. 
For stage 1 and 2, we have quadruped observations $\mathcal{O}_{\text{quadruped}}$ and the privileged observation $\mathcal{O}_{\text{privileged}}$. The action space $\mathcal{A}_{\text{quadruped}}$ is target quadruped joint angles.
% The rewards are $\mathcal{R}$, including linear velocity tracking, angular velocity tracking, quadruped base orientation, action rate, quadruped work, quadruped action norm and inward. 
For stage 3, we have the additional arm observation $\mathcal{O}_{\text{arm}}$. The action space is the union of $\mathcal{A}_{\text{quadruped}}$ and $\mathcal{A}_{\text{arm}}$ (target arm joint angles).
We train our policies in simulation using NVIDIA Isaac Gym~\cite{Rudin2021Learning} and use RMA~\cite{kumar2021rma} to transfer the policy trained in simulation to the real world. Details are on website.

\section{Experimental Setup} 
\label{sec:experimental_setup}

% \subsection{Task Setup}
In the \emph{stable locomotion task}, we learn a locomotion policy that is stable against external perturbations. 
During policy training, we sample the external pushing force uniformly from $200-300$\footnote[1]{According to the simulator's documentation, forces are in Newtons. However, we empirically found the unit to be physically unreasonable and possibly correct up to scale.} with an offset from the base sampled uniformly from $[-0.12, 0.12]m$ along the $x$ or $y$ axis of the robot quadruped base. The policy is trained to walk with a linear velocity in the range $[0,0.3]m/s$. We use a Unitree A1 with a WX200 arm.
In the \emph{agile locomotion task}, we learn a locomotion policy that can turn fast at high running speed. No external perturbations are applied in this task.
 The policy is trained to run with a linear velocity in the range $[0.6,1.5]m/s$ and an angular yaw velocity in the range $[0.5,2]m/s$. We use a Unitree Go1 with a K1 arm for this task.

In both tasks, the episode terminates when the legged robot base collides with the ground. Since the last several degrees of freedom of the arm have less effect on changing the inertia and momentum, we only use the first 3 degrees of freedom of the arm. More details are on website.

%\subsection{Metrics}
% \paragraph{Metrics} 
The following metrics are used to evaluate the results: \emph{Success rate}, the fraction of robots that survived at the end of the episode; \emph{Time to fall} (TTF), the average episode length across all the robot instances divided by the total episode length; \emph{Linear velocity ($V_{lin}$) tracking error}, the L2 distance between the robot base linear velocity and the linear velocity command; and \emph{Angular velocity ($V_{ang}$) tracking error}, the L2 distance between the robot base yaw angular velocity and the yaw angular velocity command.
\section{Quantitative Analysis: Benefit of Arm for Locomotion} \label{sec:results-stability}

\begin{table*} [h!]
\vspace{5pt}
\begin{center}
 \begin{tabular}{ c||cc cc cc}
\multirow{2}{*}{Arm} & \multicolumn{2}{c}{\textbf{Success \%} ($\uparrow$)} & \multicolumn{2}{c}{\textbf{TTF \%} ($\uparrow$) } & \multicolumn{2}{c}{\textbf{$V_{lin}$ Tracking Error} ($\downarrow$)}\\
\cline{2-7}
 & 200-300& 300-400 & 200-300& 300-400 & 200-300& 300-400\\
 \hline
No Arm  & 95.02$\pm$0.62& 68.03$\pm$1.14 & 0.975$\pm$0.00& 0.809$\pm$0.01   & 0.230$\pm$0.02& 0.765$\pm$0.03 \\
Locked  & 90.61$\pm$0.77& 45.83$\pm$1.71  & 0.950$\pm$0.00& 0.627$\pm$0.01   & 0.067$\pm$0.01& 0.224$\pm$0.01 \\
Actuated&\textbf{96.11}$\pm$0.37& \textbf{78.20}$\pm$1.77 & \textbf{0.982}$\pm$0.00&\textbf{0.875}$\pm$0.01   & \textbf{0.027}$\pm$0.00& \textbf{0.120}$\pm$0.02\\
\end{tabular}
\end{center}
\vspace{-3mm}
\caption{Results for locomotion under disturbances. 
Mean and std are computed over 5 seeds.}
\label{tab:results_set1}
\vspace{-15pt}
\end{table*}

\textbf{Stability against perturbation} We compare the performance of the robot with an actuated arm, with a locked arm ($\pi_2$), and without arm (Table~\ref{tab:results_set1}). The robot with an actuated arm outperforms both baselines, having a survival rate of up to 33 percentage points higher and a linear velocity tracking error up to $88\%$ lower. The success rate and TTF(\%) for all policies decrease as the external perturbation magnitude increases. However, the robot with an actuated arm has a relatively lower performance drop while that of a frozen arm has a 45 percentage points performance drop for the survival rate. Not having an arm increases performance compared to keeping the arm locked, but is still outperformed by our approach. This indicates the robot cannot tolerate high perturbations without the arm's support.

\begin{table} [h!]
\begin{center}
\setlength{\tabcolsep}{4pt} 
% \resizebox{\textwidth}{!}{
 \begin{tabular}{c| c|cccccc}
$V_{lin}$ & Setting & \textbf{Success} ($\uparrow$)  &  \textbf{$V_{ang}$ TE}($\downarrow$) & \textbf{$V_{lin}$ TE} ($\downarrow$) \\
 \hline
\multirow{3}{*}{1}&No Arm & \textbf{90.29}$\pm$ 0.46  & 0.332$\pm$0.023  & 0.014$\pm$0.001\\
&Locked & 82.32$\pm$0.51  & 0.234$\pm$0.012  & 0.015$\pm$0.001\\
&Actuated & 89.64$\pm$0.64  & \textbf{0.184}$\pm$0.012  & \textbf{0.010}$\pm$0.001\\
\hline
\multirow{3}{*}{1.25}&No Arm & \textbf{89.14}$\pm$0.70 & 0.300$\pm$0.019  & 0.022$\pm$0.001 \\
&Locked  & 77.81$\pm$1.01 & 0.228$\pm$0.010 & 0.025$\pm$0.001  \\
&Actuated   & 86.78$\pm$0.77 & \textbf{0.178}$\pm$0.010 & \textbf{0.015}$\pm$0.001\\
\hline
\multirow{3}{*}{1.5}&No Arm & \textbf{86.48}$\pm$0.85  & 0.291$\pm$0.017  & 0.031$\pm$0.001\\
&Locked  & 69.77$\pm$0.47 & 0.226$\pm$0.008 & 0.037$\pm$0.001 \\
&Actuated   & 82.96$\pm$0.74 & \textbf{0.175}$\pm$0.009 & \textbf{0.024}$\pm$0.001\\
\end{tabular}
% }
\end{center}
\vspace{-3mm}
\caption{Simulation evaluation for high-speed locomotion. Mean and std are computed over 5 seeds. Velocity commands are in $m/s$, success in percent, and TE denotes tracking error.
% Velocities are sampled at $1$, $1.25$ and $1.5$ $m/s$. Across all velocities, our method with the arm actuated always outperforms our method with the arm locked, across the success rate, angular velocity tracking, and linear velocity tracking. 
}
\label{tab:results_turn}
\vspace{-7mm}
\end{table}

\textbf{Agile locomotion} We compare all methods on the task of high-speed locomotion in simulation. We average the performance over 2000 trials, with a maximum duration of 20s. We sample a yaw angular command of 0.5 $rad/s$ for half of the trials and 2 $rad/s$ for the rest, and a constant linear velocity command of 1$m/s$, 1.25$m/s$, and 1.5$m/s$. 
% We evaluate the policies using the following metrics: \textit{Success rate}; \textit{Angular velocity tracking}, the L2 distance between the robot base yaw angular velocity and the yaw angular velocity command; and \textit{Linear velocity tracking}.
%
Results are shown in Table~\ref{tab:results_turn}. The policy with an actuated arm has better tracking performance than all the baselines. However, in terms of success rate, the no-arm baseline is slightly better than the actuated arm, with differences increasing as a function of speed. This shows that the arm's weight and inertia make agile locomotion more difficult. However, actuation provides a significant advantage over a locked arm.

% For all linear velocity commands, the robot with the actuated arm outperforms the robot with the locked arm, for all metrics. As the linear velocity command increases, the performance discrepancy between the actuated arm and the locked arm also increases, indicating that the arm motion especially helps the robot turn at higher speeds.

% \begin{figure}
%     \begin{minipage}[t]{0.49\textwidth}
%     \includegraphics[width=\textwidth]{figures/simanay.png}
%     \caption{Simulation results when robot is running at 1$m/s$. The desired (light green) and actual (dark green) first arm joint position are shown with the quadruped base angular velocity command (blue) and the quadruped yaw (orange).}
%     \label{fig:sim_ana}
%     \end{minipage}
% \hfill
%     \begin{minipage}[t]{0.5\textwidth}
%     \includegraphics[width=\textwidth]{figures/yaw_arm.png}
%     \caption{Simulation correlation analysis between arm motion and robot yaw. The x and y axis show the first arm joint position and the robot yaw. The blue line is a trending line fitted with linear regression on raw values (blue dots). }
%     \label{fig:sim_yaw}
%     \end{minipage}
%     \vspace{-20pt}
% \end{figure}

\begin{figure}
    \begin{subfigure}[t]{0.45\textwidth}
        \includegraphics[width=\textwidth]{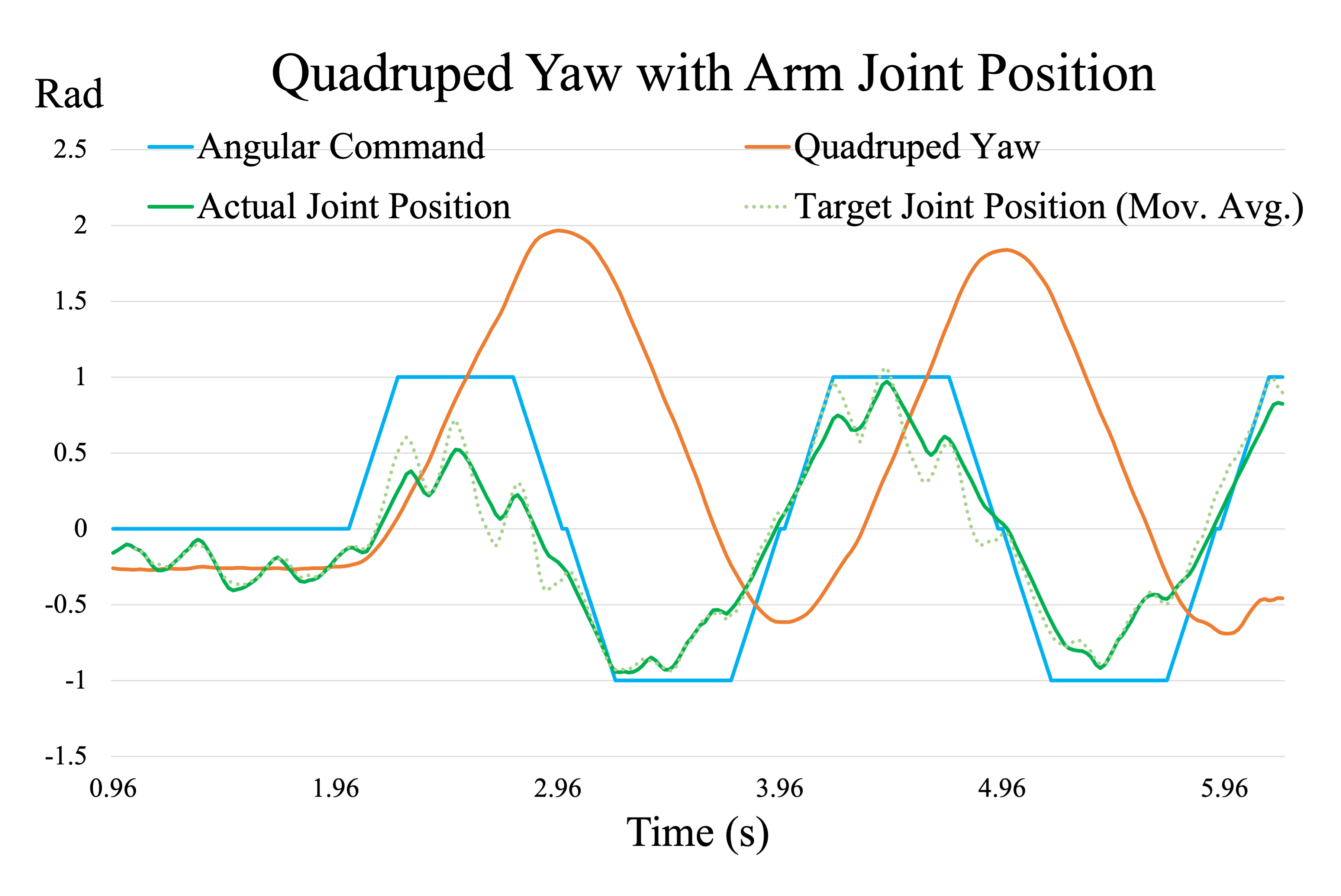}
        \vspace{-5mm}
        \caption{Simulation results with robot running at 1$m/s$. The desired (dotted green) and actual (solid green) first arm joint positions are shown with the quadruped base angular velocity command (blue) and yaw (orange).}
        \label{fig:sim_ana}
    \end{subfigure}
    \hfill
    \begin{subfigure}[t]{0.45\textwidth}
        \includegraphics[width=\textwidth]{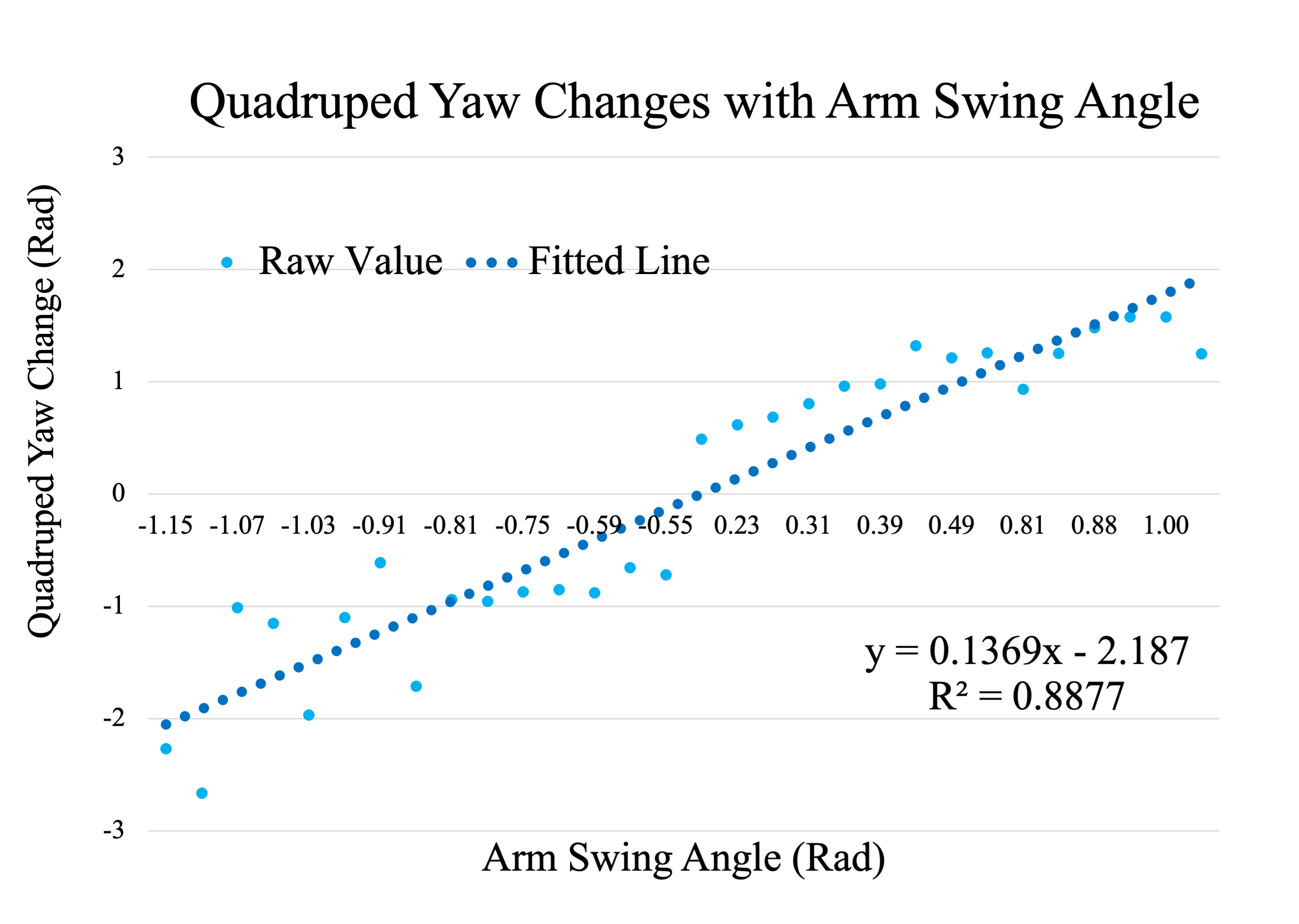}
        \caption{Simulation correlation analysis between arm motion and robot yaw. 
        % The x and y axis show the first arm joint position and the robot yaw. 
        The blue line is a trending line fitted with linear regression on raw values (blue dots).}
        \label{fig:sim_yaw}
    \end{subfigure}
    \caption{Simulation results and analysis of the benefit of arm for agile locomotion.}
    \label{fig:simulations}
    \vspace{-24pt}
\end{figure}

We conduct simulation experiments studying the relation between the arm motion and the quadruped yaw as in Fig.~\ref{fig:sim_ana}. We study the first arm joint motion as it has the largest effect on inertia. In Fig.~\ref{fig:sim_ana}, we show the first arm joint state and robot yaw for robot running at 1$m/s$ with a changing yaw angular command up to 1$rad/s$. From the plot, the arm starts to move right after receiving a non-zero yaw command and the yaw of the quadruped changes after the arm motion. The delay between the quadruped yaw and the arm motion indicates that the quadruped is following the arm motion and that the arm motion is an active choice of the policy to initiate robot turning, similar to cheetah's tail behaviour, instead of a passive behaviour caused by the inertia.

We study the relation between the magnitude of the arm motion and the quadruped base motion as in \cite{Siddall2021mechanisms}. We record the robot state for robot running at 1.5$m/s$ with a changing yaw angular velocity command. 
We sample 10 angular yaw commands and 3 points from each command to record the first arm joint position and the quadruped yaw position. Details are on website.
% We sample 10 yaw angular command from [0.5, 1.7] $rad/s$ and [-1.7,-0.5] $rad/s$ with a 0.3 $rad/s$ interval. For each trial with a different yaw command, we sample 3 points at the maximum yaw command and record the first arm joint position and the quadruped yaw position, resulting 30 pairs of points. 
We plot the quadruped yaw position with respect to the first arm joint position, shown in Fig.~\ref{fig:sim_yaw}. Linear regression indicates a strong positive correlation between the arm motion and the quadruped yaw motion with $R^2 = 0.8877$, consistent with the gecko tail behaviour in \cite{Siddall2021mechanisms}. 

\section{Quantitative Simulation Analysis}

\begin{table*} [h!]
\vspace{5pt}
\begin{center}
 \begin{tabular}{ c||cc cc cc}
\multirow{2}{*}{Policy} & \multicolumn{2}{c}{\textbf{Success \%} ($\uparrow$)} & \multicolumn{2}{c}{\textbf{TTF \%} ($\uparrow$) } & \multicolumn{2}{c}{\textbf{$V_{lin}$ Tracking Error} ($\downarrow$)}\\
\cline{2-7}
 & 200-300& 300-400 & 200-300& 300-400 & 200-300& 300-400\\
 \hline
PPO     & 64.43& 23.10  & 0.770& 0.405   & 0.028& \textbf{0.029} \\
PPO cur & 76.85& 34.40  & 0.843& 0.458   & 0.032& 0.114  \\
Fine-tuned $\pi^2$ & 69.75& 37.03  & 0.807& 0.543  &0.023& 0.066 \\
Decoupled & 92.10& 68.48  &0.930& 0.734  & 0.085& 0.222  \\
One stage demonstration & 87.93& 55.03 & 0.934& 0.692   & 0.023& 0.077 \\
Ours &\textbf{95.73}& \textbf{84.73} & \textbf{0.975}&\textbf{0.916}   & \textbf{0.021}& 0.081\\
\end{tabular}
\end{center}
\caption{Simulation evaluation of all methods for stabilizing under external pushes. 
}
\label{tab:results_set2}
\vspace{-10pt}
\end{table*}
% \begin{table*} [h!]
% \begin{center}
%  \begin{tabular}{c| c|cccccc}
% Force & Policy & \textbf{Success \%} ($\uparrow$) & \textbf{TTF \%} ($\uparrow$)  & \textbf{$V_{lin}$ Tracking} ($\downarrow$)\\
%  \hline
% \multirow{7}{*}{200-300}&PPO     & 64.43 & 0.770   & 0.028 \\
% &PPO cur & 76.85  & 0.843   & 0.032  \\
% &Locked    & 91.33 & 0.952  & 0.034 \\
% &Fine-tuned $\pi^2$ & 69.75 & 0.807  & 0.023 \\
% &Decoupled & 92.10 & 0.930   & 0.085  \\
% &One stage demonstration & 87.93 & 0.934   & 0.023 \\
% &3-stage learning& \textbf{95.73} & \textbf{0.975}   & \textbf{0.021}\\
% \hline
% \multirow{7}{*}{300-400}&PPO & 23.10 & 0.405   & 0.029 \\
% &PPO cur & 34.40  & 0.458  & \textbf{0.028}  \\
% &Locked    & 55.30 & 0.707   & 0.114 \\
% &Fine-tuned $\pi^2$& 37.03 & 0.543   & 0.066 \\
% &Decoupled & 68.48 & 0.734   & 0.222  \\
% &One stage demonstration & 55.03 & 0.692  & 0.077 \\
% &3-stage learning& \textbf{84.73} & \textbf{0.916}    & 0.081\\
% \end{tabular}
% \end{center}
% \caption{Simulation evaluation results with in training distribution pushing forces (200-300) and out of training distribution pushing forces (300-400). 
% }
% \label{tab:results_set2}
% \end{table*}
We compare our incremental stagewise learning method against five baselines of increasing complexity in the stable locomotion task setting: (i) \textit{Proximal Policy Optimization (PPO)}: a policy trained from scratch on the complete whole-body control task with PPO. (ii) \textit{PPO Curriculum}: a PPO policy trained with the curriculum of a quadruped with a locked arm with no perturbations, to that with perturbations and to that with a moving arm. (iii) \textit{Fine-tuned $\pi^2$}: a policy fine-tuned on $\pi^2$ for a quadruped with a moving arm. (iv) \textit{Decoupled}: We train the locomotion and arm policies separately. Specifically, we first train a locomotion policy with arm fixed ($\pi^2$) and then the arm policy with the latter locomotion policy fixed. At test time, we combine the output of these two policies. (v) \textit{One stage demonstration}: a policy trained with $\pi^1$ being a teacher policy on a quadruped with a moving arm directly, without multi-stage learning. 

% We evaluate all the policies using the following metrics: \textit{Success rate}, the fraction of robots that survived at the end of the episode; \emph{Time to fall} (TTF), the average episode length across all the robot instances divided by the total episode length;  and \emph{Linear velocity tracking}, the L2 distance between the robot base linear velocity and the linear velocity command.
We evaluate with pushing forces in the training distribution (200-300), and with pushing forces beyond what was seen in training (300-400). We average the performance of all methods over 2000 trials, each with a maximum duration of twenty seconds of simulated time. For 50\% of the trials, we sample a standing still command and, for the rest, a forward velocity command of 0.3 m/s. 
% We apply disturbance forces and torques of various magnitudes and directions.

\begin{figure}[]
    \centering
    \includegraphics[width=0.5\textwidth]{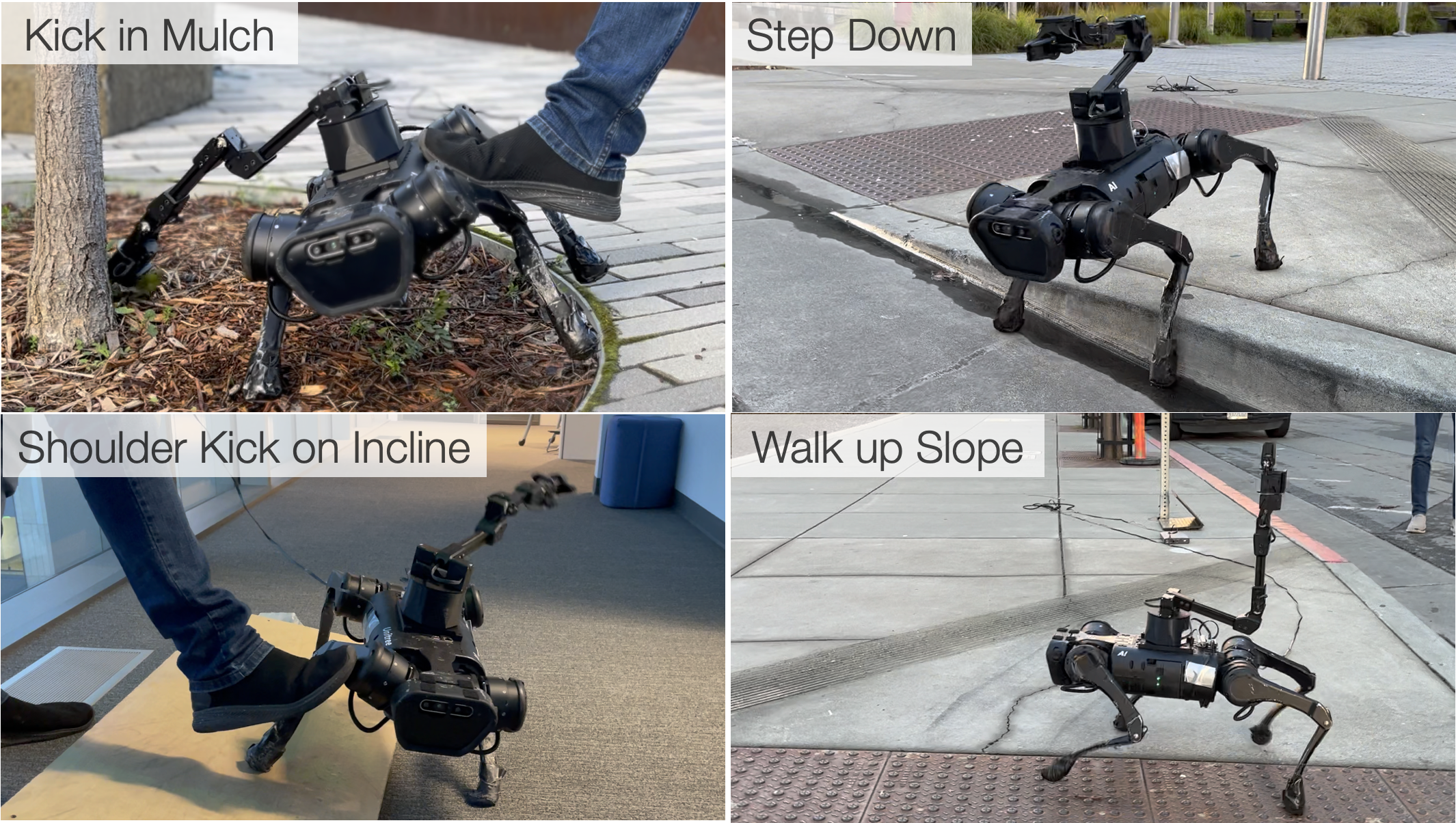}
    \caption{Physical Experiments under different scenarios. Our system, only trained on fractal terrains with external pushes, is successfully deployed on different real-world scenarios. The arm is at different configurations under different scenarios.
    %We deploy our system in various real-world scenarios as shown, none of which were seen during training. The robot has only been trained on fractal terrains with external pushes. 
    % On the incline, the arm moves to generate the counter force on the body of the quadruped to balance it. We also push from the side while it rests unevenly (sideways) on an incline, and outdoors on the muddy ground. The arm goes all the way to touch the ground for balance, but we stop since the arm hardware is not strong enough to support the quadruped. When evaluated on unleveled ground caused by lifting the leg with an incline, the robot maintains its roll roughly evenly despite the extreme perturbation. We also ask the robot to go down a step, upslope, and downslope, none of which were seen during training. We observe that while walking on the mild concrete upslope, the arm maintains a different mean position than when walking on a flat and actively uses it to get unstuck. (Video: \url{https://tinyurl.com/2p8edezu})
    }
    \label{fig:realworld}
    \vspace{-5pt}
\end{figure}

\begin{figure}
    \centering
    \includegraphics[width=0.5\textwidth]{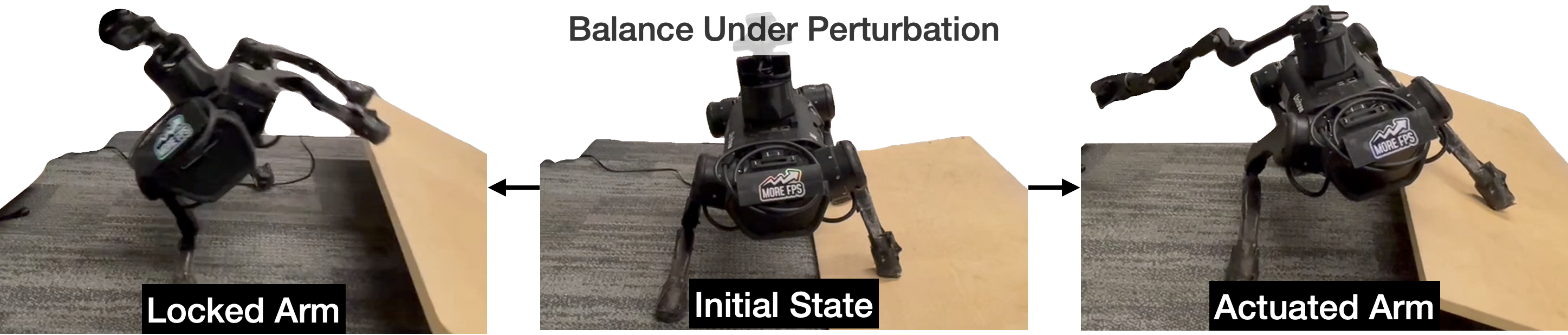}
    \caption{Robot stabilization under external perturbations. The robot with a locked arm fails to maintain balance. For actuated arm, instead of moving the arm to its left for changing the CoM, as shown in section~\ref{sec:math}, the arm relies on the dynamic torques to mitigate the impulse perturbation, resulting the arm moving to its right.}
    \label{fig:push_comp}
     \vspace{-20pt}
\end{figure}

\begin{figure*}[t!]
    \centering
    \vspace{5pt}
    \includegraphics[width=0.8\textwidth]{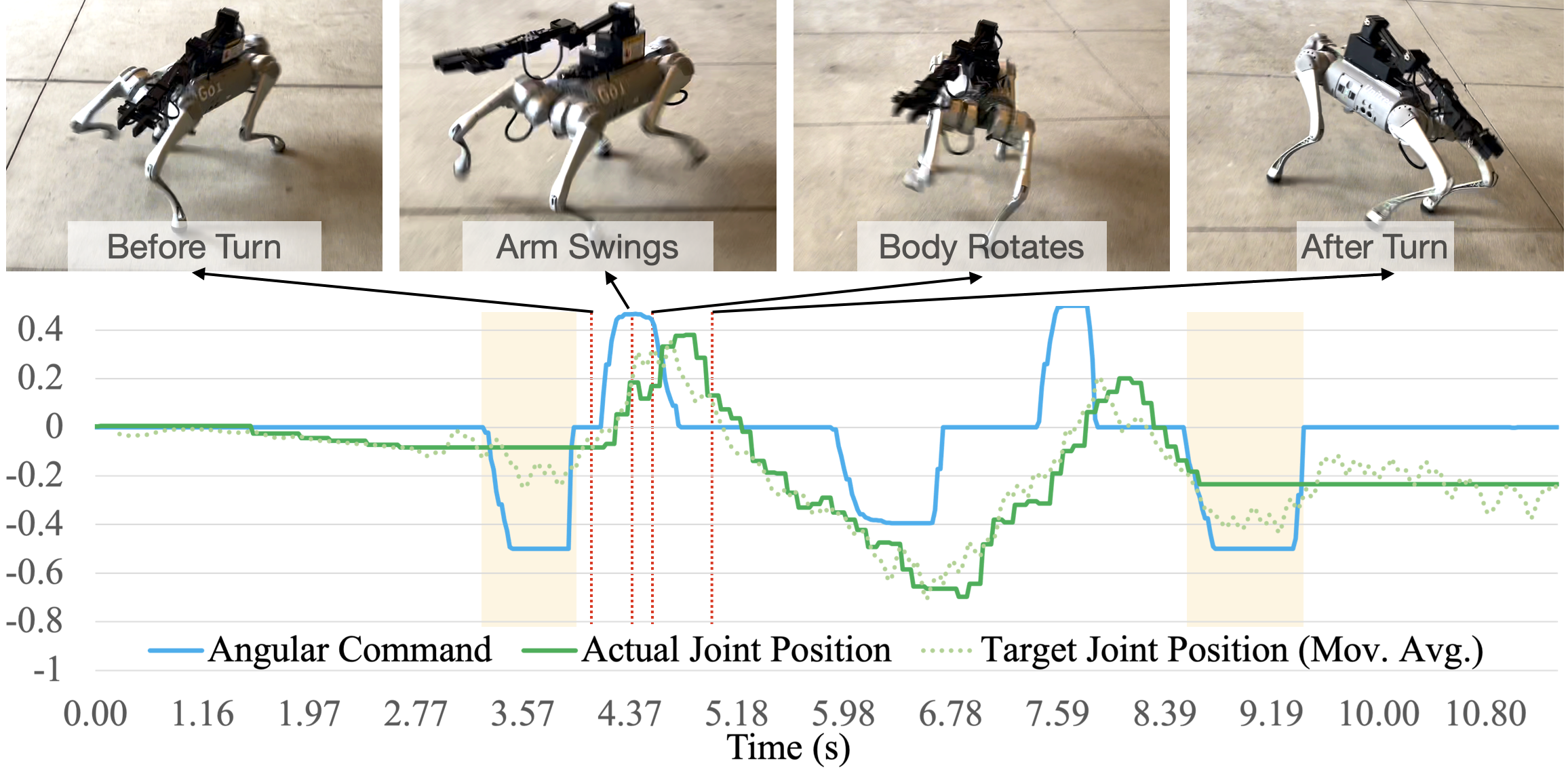}
    \caption{Physical experiments analysis. The arm anticipates the body during high-speed rotation.
    % receives the yaw command (blue). The desired first arm joint position (solid green) given by the policy and the actual position (dotted green) is shown . 
    % The video frames correspond to time steps on the plot: (1) The robot base is rotated to the right with the arm near the middle position. (2) The robot receives the command to turn left, the arm moves towards left to initiate the robot turning. (3) The robot turns, following the arm motion. (4) The robot finishes turning and the arm moves back to the middle nominal position.
    }
    \label{fig:real_ana}
    \vspace{-20pt}
\end{figure*}

Results are shown in Table~\ref{tab:results_set2}. Our method outperforms all the baselines in success rate and TTF(\%), achieving up to 16 percentage points higher survival rate than the next best method. PPO trained from scratch has the lowest performance, likely because joint training for the final task is very complex, leading to poor gradient estimates that could lead to a poor local optimum. The PPO performance falls by the largest magnitude compared to any of the other baselines for out of distribution pushing forces. This shows that vanilla PPO has a much poorer generalization. The decoupled baseline performs worse than our method. This is likely because learning a controller for the arm separately from a locomotion controller fails to yield coordinated behavior for better overall performance. This happens despite training the locomotion controller with external pushes, as in all our baselines. The baseline with incremental learning but without using demonstrations from a previous policy (``PPO cur" in Table~\ref{tab:results_set2}) performs better than vanilla PPO but significantly worse than our method. Using demonstrations but without using incremental stage-wise learning (``One stage demonstration" in Table~\ref{tab:results_set2}) shows a significant performance drop for out of distribution pushing forces. This indicates the effectiveness of our method of combining RL, BC and incremental learning. Our method shows a slight performance drop between the two push force ranges, likely because the rotational degree of freedom in the base of the arm allows it to generate a counteracting angular velocity.
For linear velocity tracking, our policy performs the best for pushing forces within distribution. Note that our policy needs to track velocity amidst more extreme perturbations compared to what the baselines experience since it has a higher survival rate. 
% Moreover, our method maintains roughly similar tracking performance between the training and out-of-distribution evaluation.

% \vspace{-1mm}
\section{Physical Experiments} 
\subsection{Locomotion under Disturbances}

We deploy our system, trained only in simulation on fractal terrains with external pushes, in a variety 
of static and dynamic real-world scenarios, some of which are shown in Fig.~\ref{fig:realworld}. We test with perturbations on an incline, mulch, slopes, and down steps, none of which were simulated during training. Our deployment results are shown in the supplementary video. 

We evaluate on an incline balanced on a fulcrum positioned in the middle. We put the robot on one end of the incline and rotate the incline by exerting downward pressure on the other end. The arm moves to generate a counter torque on the trunk of the quadruped, ensuring the robot's pitch stays roughly level during the rotation. However, at equilibrium, the arm does not move back to shift the center of mass towards the incline, since the arm is too light to have a significant impact on the center of mass (see Sec.~\ref{sec:math} for details).
We vary the incline angle irregularly, causing the arm to respond quickly by changing the direction of rotation in response to the incline rotation direction. We also create an unleveled ground by keeping the right half of the robot on a pivoting incline, while the other two legs stay on the ground. The robot can handle extreme rotations to the plank despite only having seen fractal variations to flat ground in simulation.
% We push the robot from the side while it rests unevenly (sideways) on an incline, and outdoors on muddy ground. In both cases, the arm moves to balance the robot.
% going all the way to touch the ground for balance. We observe this behavior in simulation, but the arm hardware in the real world is not strong enough to support the arm, so we stop at the limit. 
The robot can traverse a step, upslope, and downslope, none of which were seen during training. While walking on the mild concrete upslope, the arm maintains a different mean position compared to when walking on flat terrain, and helps it get unstuck while walking up a mild slope.

We additionally conduct quantitative experiments comparing the performance of a quadruped with an actuated and a locked arm. The robot is placed on a moving incline that is pushed down to create an external perturbation. When the external perturbation is high, the quadruped with a locked arm falls on the ground, while the one with a moving arm maintains stable as shown in Fig.~\ref{fig:push_comp}. The roll changes of the quadruped with the locked arm is twice as high as that of the quadruped with a moving arm (plot on website). This suggests our method of using the arm as a tail can help the quadruped to stabilize under extreme perturbations.

\subsection{Agile Locomotion}
We conduct experiments where the robot is running (1-2$m/s$) and turning (0.5-1$rad/s$) at high speeds on different terrains including grassland (Fig.~\ref{fig:fig1}) and cement ground (Fig.~\ref{fig:real_ana}). We record the the first arm joint state for robot running at 1$m/s$ and turning at different speeds in Fig~\ref{fig:real_ana}.
% Only the first arm joint state is recorded as it has the largest effect on generating the momentum.
% The plot in Figure.~\ref{fig:real_ana} shows the robot yaw angular velocity command, the desired joint position of the first arm joint and the actual joint position of the first arm joint. 
The video frames on top show the robot state at different time steps. The arm starts to move from right to left when an angular command to the left is received as shown by the first two frames. The quadruped then starts to turn following the arm motion, shown in the third frame. That the arm swings before the quadruped start to turn, indicating the arm motion initiates the quadruped motion by generating momentum, with a delay consistent of what observed in simulation (see Sec.~\ref{sec:math} for an analytical analysis of this behavior).
%
%The delay between body motion and arm motion is consistent with the simulation results. 
%The arm moves back to the nominal position after the angular command becomes $0$. 
However, due to limitations on the arm hardware, the arm can fail to achieve the action given by the policy as indicated by the orange area in the plot, where the desired arm position and the actual arm position has a high discrepancy.

\section{First Principles Analysis of Arm Motion}
\label{sec:math}
We provide an analytical derivation of the static and dynamic effects of the manipulator motion on the robot body.
To make the analysis tractable, we approximate the robot body as a cuboid and the arm as a rod attached to the robot's center of mass (CoM).
First, a static analysis shows that changes in the system's center of mass $cCoM$ due to the arm rotation $\theta$ are minimal. Indeed, under the previous assumption, the latter can be written as:
\begin{align}
    cCoM  = \frac{M_a}{M_a+M_b} \frac{l}{2} \cos\theta = 0.022 \cos \theta \leq 0.022m,
\end{align}
where $M_a, M_b$ are the arm's and body's mass, respectively, and $l$ is the arm length.
Second, a dynamic torque analysis shows that the arm's angular acceleration $\alpha_a$ can significantly affect the body's angular acceleration $\alpha_b$. Specifically, it can be shown that
\begin{align}
    \alpha_b &= -\frac{4 M_a l^2}{M_b (h^2+w^2)} \alpha_a = -0.378 \alpha_a,
    %\alpha_b &= -0.378 \alpha_a
\end{align}
where $h$ and $w$ are the torso's height and width.
This acceleration is used by our policy to absorb parts of the torques induced by disturbances or during high-speed locomotion to get faster and sharper turns, resulting in much better angular velocity tracking.
We refer to the supplementary material on the project website for a derivation and a similar analysis for changes in the system's center of pressure.

% will derive the changes in CoM derived by the arms motion and show that the rotation of the arm has relatively little influence on the CoM of the system.
% %
% Then, we will compute the torques and angular acceleration induced on the robot by the manipulator changes and show that acceleration of the arm can result in significant acceleration on the robot's body.
% %
% Overall, this section shows that the dynamic effects due to the arm's motion are much more prominent than the static ones and builds intuition upon the behavior of our policy when subject to disturbances and moving at high speed.

\section{Discussion and Limitations}
Attaching an arm to legged robots expands the legged robots capability to mobile manipulation but introduces challenges to robot stability due to increased CoM and additional payload. In this work, we mitigate this challenge by actively controlling the arm for enhanced stability and agility.
Specifically, we utilize incremental stage-wise learning to use the arm of a quadruped robot for balancing on uneven terrain, under external pushes and turning at high speeds.
Through simulation experiments and real-world deployment, we have shown quantitatively and qualitatively that an arm can be an asset for improving the stability and agility of locomotion.
Our results and analytical analysis indicate that an exciting avenue for future research is designing a light “tail” capable of fast accelerations.
However, our policy is blind and does not use the environment's affordances to balance. Using vision and real-world experience to improve could potentially alleviate this problem~\cite{loquercio2022learning}. The proposed incremental stage-wise learning approach requires manually designing each stage, which could be challenging for more complex systems. Extending the current approach to automatically designed stages could be an interesting future direction.

% does not use environment affordances to help balance. -> Use this sentence in camera ready since it is more juicy

% One limitation of the proposed approach is that the policy was only trained in simulation and does not use real-world experience to improve. One way of doing this effectively is using the coupling between different sensors and actuators~\cite{loquercio2022learning}. In addition, the manipulator arm is attached to the robot, we don't execute manipulation tasks. Using the arm to balance the robot while simultaneously executing manipulation tasks is an interesting avenue for future work. Our performance is also limited by the arm hardware. It would be interesting to use a more capable arm in the future.

% \section{Conclusion} 
% \label{sec:conclusion}

% We believe that our approach could be used to simplify the optimization of many other problems beyond locomotion that vanilla reinforcement learning cannot sufficiently tackle.
%Using Chimera, we learned a controller that enables a quadruped robot with an arm to carry out locomotion, staying balanced against external forces. 
% Applying our algorithm to tasks beyond whole-body control is an interesting venue for future work.
%

% Through simulation experiments and real-world deployment, we have shown quantitatively and qualitatively that our policy successfully carries out stable locomotion, using the arms and legs in synergy to stay balanced. 

\section*{Acknowledgements}
\footnotesize
This research was performed at UC Berkeley in affiliation with the Berkeley AI Research (BAIR) Lab, and the CITRIS "People and Robots" (CPAR) Initiative. The authors were supported in part by donations from Toyota Research Institute.
% \input{RSS_citation_info.tex}

% \renewcommand*{\bibfont}{\footnotesize}
% % \renewcommand{\baselinestretch}{0.9} %only do this for refs if crunched for space
% \printbibliography
% \clearpage
% \section*{Acknowledgments}

%% Use plainnat to work nicely with natbib. 

% \bibliographystyle{plainnat}
\bibliographystyle{IEEEtran}
\bibliography{IEEEabrv,references}

\end{document}